%% file: main.tex
\documentclass[conference]{IEEEtran}
\IEEEoverridecommandlockouts
\usepackage{cite}
\usepackage{amsmath,amssymb,amsfonts}
\usepackage{algorithm,algorithmic}
\usepackage{graphicx}
\usepackage{textcomp}
\usepackage{xcolor}
\def\BibTeX{{\rm B\kern-.05em{\sc i\kern-.025em b}\kern-.08em
    T\kern-.1667em\lower.7ex\hbox{E}\kern-.125emX}}

\usepackage{booktabs,tabularx}
\usepackage{array}
\newcolumntype{Y}{>{\centering\arraybackslash}X}
\usepackage{mathtools}
\usepackage{colortbl}
\usepackage{url,hyperref}
\usepackage{lipsum}
\usepackage{dirtytalk}
\usepackage{placeins}

\newcommand{\texthl}[2]{%
  \begingroup
  \setlength{\fboxsep}{1pt}%
  \colorbox{#1!25}{#2}%
  \endgroup
}

\begin{document}

\title{\raisebox{-0.2em}{\includegraphics[height=1.1em]{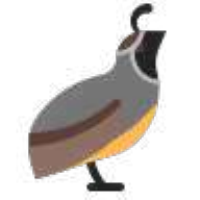}}%
\, QuAIL: Quality-Aware Inertial Learning for Robust Training under Data Corruption}


\author{\IEEEauthorblockN{Mattia Sabella\IEEEauthorrefmark{1}}
\IEEEauthorblockA{\textit{DEIB} \\
\textit{Politecnico di Milano}\\
Milan, Italy \\
\small{mattia.sabella@polimi.it}}
\and
\IEEEauthorblockN{Alberto Archetti\IEEEauthorrefmark{1}}
\IEEEcompsocitemizethanks{\IEEEcompsocthanksitem\IEEEauthorrefmark{1}Equal contribution.}
\IEEEauthorblockA{\textit{DEIB} \\
\textit{Politecnico di Milano}\\
Milan, Italy \\
\small{alberto.archetti@polimi.it}}
\and
\IEEEauthorblockN{Pietro Pinoli}
\IEEEauthorblockA{\textit{DEIB} \\
\textit{Politecnico di Milano}\\
Milan, Italy \\
\small{pietro.pinoli@polimi.it}}
\and
\IEEEauthorblockN{Matteo Matteucci}
\IEEEauthorblockA{\textit{DEIB} \\
\textit{Politecnico di Milano}\\
Milan, Italy \\
\small{matteo.matteucci@polimi.it}}
\and
\IEEEauthorblockN{Cinzia Cappiello}
\IEEEauthorblockA{\textit{DEIB} \\
\textit{Politecnico di Milano}\\
Milan, Italy \\
\small{cinzia.cappiello@polimi.it}}
}

\maketitle

\begin{abstract}
Tabular machine learning systems are frequently trained on data affected by non-uniform corruption, including noisy measurements, missing entries, and feature-specific biases. In practice, these defects are often documented only through column-level reliability indicators rather than instance-wise quality annotations, limiting the applicability of many robustness and cleaning techniques. We present QuAIL, a quality-informed training mechanism that incorporates feature reliability priors directly into the learning process. QuAIL augments existing models with a learnable feature-modulation layer whose updates are selectively constrained by a quality-dependent proximal regularizer, thereby inducing controlled adaptation across features of varying trustworthiness. This stabilizes optimization under structured corruption without explicit data repair or sample-level reweighting. Empirical evaluation across 50 classification and regression datasets demonstrates that QuAIL consistently improves average performance over neural baselines under both random and value-dependent corruption, with especially robust behavior in low-data and systematically biased settings. These results suggest that incorporating feature reliability information directly into optimization dynamics is a practical and effective approach for resilient tabular learning.
\end{abstract}


\section{Introduction}

Modern ML pipelines increasingly operate on heterogeneous tabular and sensor-derived data where quality defects are the norm rather than the exception, and recent surveys emphasize that these issues dominate the practical effort required to deploy models in production settings ~\cite{whang2023data,ogrizovic2024qualityassurance}. This shift aligns with data-centric AI, which argues that improving dataset quality can be as impactful as architectural tuning, and that model performance metrics can often act as a proxy signal for data quality interventions ~\cite{jakubik2024data}. Data errors in real-world datasets span a wide range of complexity, from explicit and readily detectable defects to subtle failures arising from drifting generation processes, biased sampling, delayed updates, or correlated source errors ~\cite{liu2024navigating_data_corruption,drift_resilient_tabpfn2024}. These latter forms of corruption often induce structured, feature-dependent distortions that are neither independent nor uniformly distributed across instances ~\cite{stress_testing_ml2025,robust_hybrid_noise2024}. Moreover, the impact of such errors on learning dynamics is frequently non-linear: small perturbations in critical or polluted features may disproportionately and unpredictably affect downstream predictions across many standard algorithms, depending on which pipeline stage and quality dimensions are compromised, while larger errors in less informative dimensions may remain comparatively benign ~\cite{mohammed2025effects,ren2023review}. As a result, the relationship between data corruption and model degradation is complex and multivariate, and cannot be adequately captured by simplistic assumptions such as uniform noise or linear performance degradation. Despite progress in data quality management, separated pre-processing–based approaches are costly to operationalize, brittle under pipeline evolution, and often assume instance-level quality labels rather than coarse column-level metadata (e.g., provenance, component reliability, freshness, or completeness) rather than instance-level quality labels ~\cite{whang2023data,jakubik2024data}. Motivated by this gap, we propose QuAIL, a quality-aware inertial learning mechanism that incorporates feature-level quality priors directly into the optimization process. QuAIL employs a learnable gating layer to modulate feature contributions and a proximal anchor regularizer that selectively slows gate updates for low-quality features, introducing quality-dependent resistance to parameter drift. As a result, reliable features adapt freely, whereas unreliable ones are constrained, thereby stabilizing training under structured corruption without data cleaning or sample-wise reweighting. Across 50 classification and regression benchmarks under random and value-dependent corruption, QuAIL consistently achieves robust performance gains over standard neural baselines and curriculum-based methods, demonstrating that treating data quality as a first-class component of optimization is an effective strategy for resilient tabular learning. The paper is structured as follows. \textit{Section II} reviews related work on data quality, robustness, and feature corruption in tabular learning. \textit{Section III} introduces the QuAIL method and its quality-aware optimization mechanism. \textit{Section IV} describes the experimental setup and corruption protocols, followed by empirical results and analysis in \textit{Section V}. Finally, \textit{Section VI} concludes the paper and outlines directions for future work.

\section{Related Work}
Data quality is commonly defined as \emph{fitness-for-use} across a dataset lifecycle \cite{DBLP:journals/jmis/WangS96}. Data quality issues (e.g., inaccuracy, incompleteness, inconsistency, and timeliness/freshness) are not merely descriptive properties but measurable risk factors for downstream learning and deployment; recent surveys introduce these dimensions, metrics, and assessment workflows specifically in the context of ML datasets \cite{gong2023datasetqualitysurvey, ehrlinger2022survey}. In tabular settings, where heterogeneous numeric/categorical columns are routinely produced by sensors, forms, and ETL pipelines, quality defects are often structured (column-dependent, pipeline-stage dependent, and sometimes value-dependent), and large-scale empirical evidence confirms that polluting data along multiple quality dimensions can yield non-linear and model-specific degradation patterns rather than uniform accuracy loss, motivating mechanisms that explicitly account for where and how corruption manifests \cite{mohammed2025effects,verde2021exploring}. Crucially, the literature emphasizes the distinction between Missing Completely At Random (MCAR), where corruption is independent of feature values, Missing At Random (MAR), where corruption may depend on observed feature values but not on the missing values themselves, and Missing Not At Random (MNAR), where errors and missingness depend directly on the unobserved data-generating process; the last two better reflects real-world dynamics and logging failures, yet is significantly harder to detect and mitigate due to systematic, feature- and value-dependent biases \cite{little2019statistical}.  A traditional solution is to isolate data quality management as a pre-processing stage (i.e., data preparation pipeline) applying profiling, repair, imputation, outlier handling, entity matching steps \cite{fan2015data}; yet data engineering highlights that cleaning is a broad family of activities with heterogeneous assumptions and repeated operational cost, and that it interacts tightly with model selection and evaluation \cite{SancriccaSC24}, making \say{clean-then-train} a simplistic approach under evolving pipelines \cite{cote2024datacleaningml}. In parallel, ML quality assurance work emphasizes that robustness must be treated as a first-class requirement alongside accuracy, fairness, and safety, and that testing/validation needs to align with the pipeline stages where faults arise \cite{ogrizovic2024qualityassurance}. These observations have fueled robust training lines that reduce reliance on explicit cleaning by modifying objectives or sampling to limit overfitting to unreliable supervision and data: the label-noise literature is particularly mature, with recent comprehensive surveys organizing strategies such as sample selection, reweighting, label correction, semi-supervised bootstrapping, and robust losses \cite{song2023noisylabels}. Curriculum learning provides a principled abstraction for these methods by ordering or weighting examples from \say{easy/clean} to \say{hard/noisy}, and modern taxonomies formalize its key design degrees of freedom, clarifying why curriculum-based baselines are appropriate whenever the training set contains a spectrum of corruption levels \cite{soviany2022curriculum}. However, while noisy-label robustness is well studied, robustness to feature corruption is less consistently addressed, and recent work on hybrid noise explicitly argues for unified formulations that treat feature and label corruption jointly via recovery-augmented objectives (e.g., low-rank/structured error modeling solved with alternating optimization) \cite{wei2024hybridnoise}. Advances in deep tabular learning increasingly rely on representation mechanisms to learn feature saliency and interactions from data, and recent surveys of self-supervised learning for non-sequential tabular data categorize these approaches and discuss how feature corruption can act as an inductive signal for robust representations \cite{wang2025ssl4tabular}. Yet, across both supervised and self-supervised tabular models, feature importance is typically inferred inherently from the observed (potentially corrupted) data distribution, whereas many real deployments also have exogenous estimated column-level quality metadata. QuAIL addresses this gap by injecting feature-quality priors directly into optimization via a learnable gating layer coupled with quality-weighted proximal anchoring, yielding controlled gate dynamics that preferentially constrain low-reliability features while preserving adaptivity for high-reliability ones.








\section{Method}

QuAIL is a simple add-on mechanism that can be integrated into any differentiable architecture for tabular data. The core idea is to introduce a learnable feature-wise gating layer that adaptively scales input features based on their reliability. High-quality features are allowed to adapt freely, whereas low-quality features are constrained by quality-aware regularization to prevent overfitting to unreliable data.

Let $f_{\boldsymbol{\theta}}: \mathbb{R}^D \rightarrow \mathcal{Y}$ denote a base prediction model with parameters $\boldsymbol{\theta}$ that maps a $D$-dimensional input $\mathbf{x}$ to an output $y \in \mathcal{Y}$. QuAIL augments this base model with a diagonal gating layer parameterized by a learnable weight vector $\mathbf{g} = [g_1, \ldots, g_D]^\top \in \mathbb{R}^D$. For a given input $\mathbf{x}_i$, the gated representation is computed as
\begin{equation}
    \tilde{\mathbf{x}}_i = \mathbf{g} \odot \mathbf{x}_i = [g_1 x_{i,1}, \ldots, g_D x_{i,D}]^\top,
\end{equation}
where $\odot$ denotes element-wise multiplication. The gated input $\tilde{\mathbf{x}}_i$ is then fed to the base model to produce the final prediction $\hat{y}_i = f_{\boldsymbol{\theta}}(\tilde{\mathbf{x}}_i)$. Each gate weight $g_j$ acts as a learned scaling factor for feature $j$, enabling the model to attenuate or amplify features based on their empirical utility for the task.

To guide the gating mechanism toward reliable features, we incorporate prior knowledge about feature quality through a proximal regularization term. Let $\mathbf{q} = [q_1, \ldots, q_D]^\top$ denote a vector of feature quality scores, where $q_j \in [0, 1]$ represents the reliability of feature $j$ (higher values indicate higher quality). We regularize the gate weights toward an anchor vector $\mathbf{g}_{\text{anchor}}$ using quality-dependent penalties:
\begin{equation}
    \mathcal{L}_{\text{gate}} = \frac{1}{D} \sum_{j=1}^{D} w_j \left(g_j - g_{j,\text{anchor}}\right)^2,
\end{equation}
where the regularization weights $w_j$ are derived from the quality scores as $w_j = \phi(1 - q_j)$. Here, $\phi: [0,1] \rightarrow \mathbb{R}_+$ is a monotonically increasing weighting function. This design ensures that high-quality features (large $q_j$) receive small weights $w_j$, allowing their gates to adapt freely, while low-quality features (small $q_j$) receive large weights $w_j$, constraining them near the anchor. In our experiments, we consider four choices for the weighting function: linear $\phi(z) = z$, quadratic $\phi(z) = z^2$, exponential $\phi(z) = e^{2z} - 1$, and inverse exponential $\phi(z) = 1 - e^{-2z}$. This quality-aware regularization implements a form of adaptive inertia: reliable features are allowed to adapt quickly to the task at hand, while unreliable features are constrained to prevent overfitting to noise.

The model is trained by minimizing a composite loss function that combines task-specific supervision with gate regularization:
\begin{equation}
    \mathcal{L} = \frac{1}{N}\sum\limits_{i=1}^N\mathcal{L}_{\text{task}}\left(\hat{y}_i, y_i\right) + \lambda \cdot \mathcal{L}_{\text{gate}},
\end{equation}
where $\mathcal{L}_{\text{task}}$ is a standard supervised loss (cross-entropy for classification, mean squared error for regression) and $\lambda > 0$ controls the strength of regularization. Both the base model parameters $\boldsymbol{\theta}$ and the gate weights $\mathbf{g}$ are optimized jointly via gradient descent.

The gate weights $\mathbf{g}$ can be randomly initialized or initialized to the quality vector directly ($g_j = q_j$) to reflect prior beliefs about feature quality. The anchor vector $\mathbf{g}_{\text{anchor}}$ is initialized identically to $\mathbf{g}$ and updated during training: every $T_{\text{anchor}}$ epochs, we set $\mathbf{g}_{\text{anchor}} \leftarrow \mathbf{g}$, allowing the regularization target to track the evolving gate configuration. This creates a moving proximal anchor that balances stability (through regularization) with adaptivity (through periodic updates). The regularization coefficient $\lambda$ can optionally be annealed during training to allow greater flexibility in the final part of training. In this work, we explore linear decay with $\lambda = \lambda_0(1 - t/T)$ and cosine decay with $\lambda = \lambda_0 \cdot (1 + \cos(\pi t / T))/2$, where $t$ is the current epoch and $T$ is the total number of training epochs.

\section{Experiments}

\begin{figure*}[t]
    \centering
    \includegraphics[width=\linewidth]{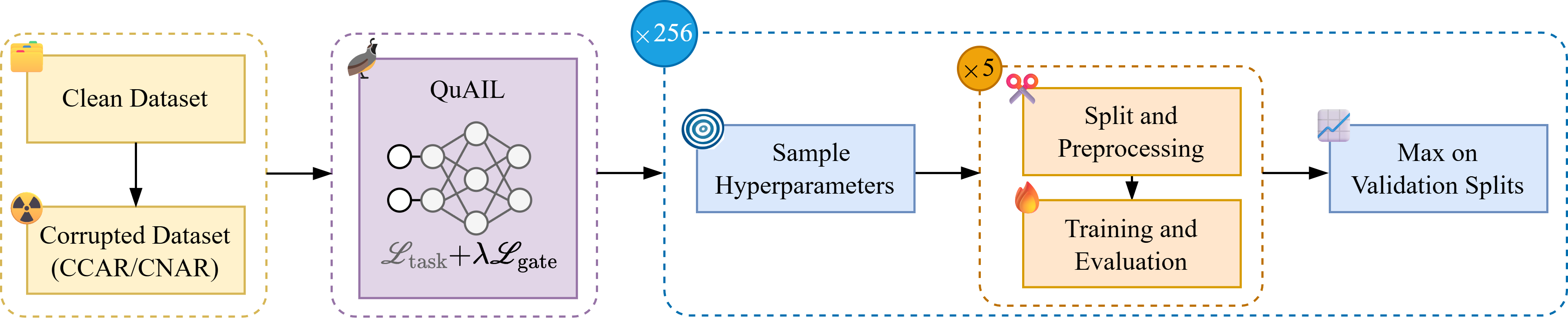}
    \caption{Optuna-based training and evaluation pipeline for QuAIL.}
    \label{fig:pipeline}
\end{figure*}

\subsection{Datasets and Data Corruption}

We evaluate our approach on 50 classification and regression datasets from the \href{https://archive.ics.uci.edu/}{UCI Machine Learning repository}~\cite{uci}, selected based on popularity and having at least 100 samples with a single supervised target. To simulate real-world data quality challenges, we introduce controlled corruption patterns under two distinct modalities: Corruption Completely At Random (CCAR) and Corruption Not At Random (CNAR). We adopt the CCAR and CNAR abbreviations by analogy with the MCAR/MAR/MNAR taxonomy for missing data~\cite{little2019statistical}, extending them to general feature corruption (comprising both missing and noisy values), with CNAR encompassing both MAR-like correlated corruption and MNAR-like value-dependent mechanisms. In all experiments, target labels remain uncorrupted to preserve the objectives of supervised learning.

Corruption severity is stratified across features. Each dataset has 6.25\% of features assigned to severe corruption (40\% of cells affected), 12.5\% to heavy corruption (20\%), 25\% to moderate corruption (10\%), and the remaining features to mild corruption (5\%). This distribution yields approximately 10\% overall cell-level corruption per dataset. Corruption is applied exclusively to the training data, ensuring that the validation and test sets remain clean for unbiased comparisons.

\subsubsection{Corruption Completely At Random (CCAR)} 

The CCAR mode simulates independent random corruption events with uniform probability across feature values. For numerical features $x_j \in \mathbb{R}$, we apply additive Gaussian noise $\tilde{x}_j = x_j + \epsilon$ with $\epsilon \sim \mathcal{N}(0, \sigma^2)$, where $\sigma = \sigma_{j} / 10$ ensures a signal-to-noise ratio of 20 dB across features of different scales. For categorical features with domain $\mathcal{C} = \{c_1, \ldots, c_k\}$, corrupted values are uniformly replaced by an alternative category: $\tilde{c} \sim \text{Uniform}(\mathcal{C} \setminus \{c\})$. Missing values are further introduced by setting 30\% of each column's corrupted cells to NaN.

\subsubsection{Corruption Not At Random (CNAR)} 

The CNAR mode implements value-dependent corruption, creating systematic biases that are more realistic than failures in CCAR. For numerical features, we apply heteroscedastic noise, in which the corruption magnitude increases with the feature value. Given a min-max normalized feature $x_j \in [0,1]$, we compute $\tilde{x}_j = x_j + (1 + 3x_j) \cdot \epsilon$ with $\epsilon \sim \mathcal{N}(0, \sigma^2)$ and $\sigma = \sigma_{j} / 10^{0.75}$ (SNR = 15 dB). Missing values follow a MNAR pattern where extreme values are preferentially deleted: $P(\text{missing} \mid x) = \min(0.5, |x - \text{median}| \cdot p_{\text{corruption}} / (2 \cdot \text{IQR}))$, with IQR denoting the interquartile range. Additionally, to model correlated measurement failures similar to MAR, we introduce corruption propagation across numerical features. One-third of numerical features are designated as anchors, and when an anchor is corrupted at row index $i$, other numerical features from $i$ have a 60\% probability of also being corrupted with independent Gaussian noise at SNR = 15 dB.

Categorical features undergo systematic confusion rather than random replacement. Each category $c_i$ is mapped to a fixed alternative $c_{(i+1) \bmod k}$, and corrupted instances follow this mapping with 70\% probability, falling back to uniform random replacement otherwise. This mirrors scenarios in which specific categories are consistently mislabeled. Missing values occur more often for rare categories as $P(\text{missing} \mid c) = p_{\text{corruption}} \cdot (1 - f(c))$, where $f(c)$ is the empirical frequency of category $c$.

\subsection{Training and Evaluation Pipeline}

We compare QuAIL against three baselines: a linear model, a standard multi-layer perceptron (MLP), and an MLP with curriculum learning. The curriculum learning baseline implements sample reweighting via a probabilistic data sampler that constructs batches with probability proportional to each sample's cleanliness, gradually exposing the model to increasingly corrupted examples during training. Since corruption is synthetically injected, sample cleanliness is defined as the fraction of uncorrupted feature values.

Our evaluation pipeline is designed to minimize overfitting on small-scale tabular datasets through bootstrap averaging, as illustrated in Figure~\ref{fig:pipeline}. For each combination of dataset, corruption mode (clean, CCAR, or CNAR), and model, we perform the following procedure. We start by sampling a hyperparameter set using a Tree-structured Parzen Estimator~\cite{ozaki2020multiobjective} provided by the Optuna framework~\cite{akiba2019optuna}. This balances hyperparameter exploration and exploitation more effectively than random search while avoiding the computational cost of grid search. Then, given a set of hyperparameters, we generate 5 bootstrap splits with 60-20-20 train-validation-test proportions. Corruption is applied only to the training data, leaving the validation and test sets clean for consistent baseline comparisons. Each split undergoes median imputation for numerical missing values and mode imputation for categorical missing values independently, followed by standardization and one-hot encoding.

For each split, we train a separate model for up to 256 epochs, with early stopping based on validation performance, using a patience of 16 epochs. The validation metric of choice is the F1-score for classification and the R$^2$ for regression. After training all 5 models from the bootstrap splits, we compute the average test performance. This average constitutes the reported result, ensuring robustness to the inherent data variability in small-scale tabular datasets.

The hyperparameter search space includes learning rate $\in [5 \times 10^{-4}, 5 \times 10^{-2}]$ (log-scale), batch size $\in \{64, 128, 256\}$, optimizer (Adam, AdamW, SGD), weight decay $\lambda_{\text{WD}} \in [10^{-6}, 10^{-2}]$ (log-scale), and learning rate scheduler (on-plateau, cosine annealing, step-based decay). For MLP models, we additionally search over the number of hidden layers (1 to 4), layer width $\{4, 8, 16, 32, 64\}$, dropout rate $\in [0, 0.5]$, and activation function (ReLU, ELU, GELU). Curriculum scheduling can follow a linear, exponential, or step-based decay. QuAIL-specific hyperparameters include gate initialization strategy (quality vector, random, all ones), regularization weight $\lambda \in [10^{-4}, 10^{-1}]$ (log-scale), anchor update frequency $T_{\text{anchor}} \in [1, 20]$ epochs, quality weighting function $\phi$ (linear, quadratic, exponential, inverse exponential), and regularization annealing schedule (constant, linear decay, cosine decay). Finally, QuAIL can optionally include curriculum in the data sampler.

We allocate 256 Optuna trials to optimize the baseline model. For QuAIL experiments, we restrict the architectural search to the top-8 performing MLP configurations from the baseline experiments and sample only QuAIL-specific hyperparameters, thereby reducing search complexity.

\subsection{Results}

\input{tab_cls}
\input{tab_reg}
\input{tab_comp}

Tables~\ref{tab:classification_results} and \ref{tab:regression_results} collect our results on classification and regression datasets, respectively. QuAIL demonstrates consistent improvements over baseline methods under both CCAR and CNAR modes, with higher gains on regression tasks under CCAR. Since small datasets may be highly sensitive to feature corruption, we use a 10\% trimmed mean when computing average relative improvements to reduce the influence of pathological datasets that exhibit degenerate behavior. Across classification datasets, QuAIL achieves average relative improvements of $+0.99$\% (CCAR) and $+1.90$\% (CNAR) in F1-score compared to the best-performing MLP baseline, while regression tasks exhibit gains of $+2.99$\% (CCAR) and $+1.30$\% (CNAR) in R$^2$.

The curriculum learning baseline shows mixed results, occasionally matching or slightly exceeding QuAIL on specific datasets but failing to achieve consistent improvements across the benchmarks. This suggests that sample-level resampling alone may be insufficient when corruption is feature-dependent rather than instance-dependent.

To better understand when QuAIL provides the greatest improvement, Table~\ref{tab:comparison} stratifies the average relative improvement over MLP across three dimensions: dataset size, feature dimensionality, and task linearity. We partition datasets into small ($N < 1{,}000$), medium, and large ($N > 10{,}000$) size categories; low ($D < 12$), medium, and high ($D > 16$) dimensionality categories; and mostly linear ($\Delta < 1.3$), mixed, and mostly non-linear ($\Delta > 8.3$) categories based on the performance gap $\Delta$ between MLP and Linear baselines on clean data. These thresholds are defined based on the tertiles of the empirical distribution across our 50 datasets, ensuring approximately balanced representation within each category. QuAIL exhibits the greatest improvements on small datasets both in terms of samples and features ($+2.95$\% and $+2.95$\% respectively), with diminishing gains as dataset size increases, aligning with the intuition that in low-data settings even limited corruptions can compromise critical portions of the training set. While present for both samples and features, this trend is more pronounced in the number of rows than in the number of columns. Finally, QuAIL shows consistent improvements across tasks with linear and non-linear decision boundaries, even when corruption is likely to disrupt the task's separability structure.

\section{Conclusions}

This paper presented QuAIL, a quality-aware inertial learning mechanism that improves robustness in tabular machine learning under feature-level data corruption. By integrating column-level quality priors directly into optimization through a learnable gating layer and quality-weighted proximal anchoring, QuAIL departs from traditional clean-then-train pipelines and sample-centric robustness methods. The proposed design promotes adaptive learning on reliable features while constraining unstable updates on unreliable ones, without requiring explicit data cleaning or instance-level quality labels.

Extensive experiments on 50 UCI classification and regression datasets show that QuAIL consistently outperforms strong MLP and curriculum learning baselines under both CCAR and CNAR corruption. Performance gains are particularly evident in small datasets, where feature-level corruption can severely distort the information available. These results indicate that sample reweighting alone is often insufficient when corruption is primarily feature-dependent.

Beyond empirical improvements, QuAIL offers practical advantages for real-world deployment: it leverages readily available feature-level quality metadata, introduces minimal architectural overhead, and integrates seamlessly into existing training pipelines, simplifying data preparation costs. Future work will explore broader corruption patterns, alternative quality-to-regularization mappings, and extensions to additional data modalities and architectures. Overall, QuAIL demonstrates that explicitly modeling feature reliability in the optimization process is an effective and scalable strategy for robust learning under realistic data-quality constraints.

\bibliographystyle{IEEEtran}
\bibliography{references}

\end{document}

%% file: tab_cls.tex
\begin{table*}[t]
\caption{Test F1 Score average over 5 bootstrap executions for each classification dataset. The best model is highlighted in \texthl{green}{\textbf{bold}}.}
\label{tab:classification_results}
\begin{center}
\begin{small}
\begin{sc}
\begin{tabularx}{\textwidth}{@{} l c | Y Y | Y Y Y Y | Y Y Y Y}
\toprule
Model; Dataset (\href{https://archive.ics.uci.edu/}{UCI ID}) & Shape & \multicolumn{2}{c|}{Clean} & \multicolumn{4}{c|}{CCAR} & \multicolumn{4}{c}{CNAR} \\
\midrule
Linear &  & $\checkmark$ &  & $\checkmark$ &  &  &  & $\checkmark$ &  &  &  \\
MLP &  &  & $\checkmark$ &  & $\checkmark$ & $\checkmark$ & $\checkmark$ &  & $\checkmark$ & $\checkmark$ & $\checkmark$ \\
Curriculum &  &  &  &  &  & $\checkmark$ & ($\checkmark$) &  &  & $\checkmark$ & ($\checkmark$) \\
QuAIL &  &  &  &  &  &  & $\checkmark$ &  &  &  & $\checkmark$ \\
\midrule
Adult (\href{https://archive.ics.uci.edu/dataset/2}{2}) & $48842 \times 14$ & $35.83$ & $36.86$ & $35.29$ & $36.87$ & $36.32$ & \cellcolor{green!25}$\mathbf{37.73}$ & $35.11$ & $36.54$ & $37.09$ & \cellcolor{green!25}$\mathbf{37.42}$ \\
AIDS (\href{https://archive.ics.uci.edu/dataset/890}{890}) & $2139 \times 23$ & $73.23$ & $79.34$ & $70.29$ & $74.16$ & $73.38$ & \cellcolor{green!25}$\mathbf{74.36}$ & $69.47$ & $74.85$ & $74.45$ & \cellcolor{green!25}$\mathbf{75.35}$ \\
Bank Marketing (\href{https://archive.ics.uci.edu/dataset/222}{222}) & $45211 \times 16$ & $45.91$ & $50.85$ & $45.75$ & \cellcolor{green!25}$\mathbf{48.91}$ & $44.40$ & $46.82$ & $44.41$ & $50.81$ & $50.11$ & \cellcolor{green!25}$\mathbf{51.02}$ \\
Breast Cancer (\href{https://archive.ics.uci.edu/dataset/14}{14}) & $286 \times 9$ & $39.12$ & $45.02$ & $37.40$ & $40.19$ & $36.73$ & \cellcolor{green!25}$\mathbf{50.49}$ & $39.65$ & $42.22$ & $32.46$ & \cellcolor{green!25}$\mathbf{48.88}$ \\
Breast Cancer WD (\href{https://archive.ics.uci.edu/dataset/17}{17}) & $569 \times 30$ & $96.17$ & $96.12$ & $95.43$ & $95.73$ & \cellcolor{green!25}$\mathbf{96.13}$ & $95.68$ & $95.08$ & $96.49$ & $96.38$ & \cellcolor{green!25}$\mathbf{96.68}$ \\
Breast Cancer WO (\href{https://archive.ics.uci.edu/dataset/15}{15}) & $699 \times 9$ & $95.43$ & $94.01$ & $95.21$ & \cellcolor{green!25}$\mathbf{95.62}$ & $95.23$ & \cellcolor{green!25}$\mathbf{95.62}$ & $95.27$ & $95.33$ & $95.26$ & \cellcolor{green!25}$\mathbf{95.71}$ \\
Car Evaluation (\href{https://archive.ics.uci.edu/dataset/19}{19}) & $1728 \times 6$ & $86.33$ & $98.23$ & $65.50$ & $66.27$ & \cellcolor{green!25}$\mathbf{68.72}$ & $68.16$ & $60.40$ & $67.40$ & $69.44$ & \cellcolor{green!25}$\mathbf{76.06}$ \\
Cirrhosis (\href{https://archive.ics.uci.edu/dataset/878}{878}) & $418 \times 17$ & $52.62$ & $62.17$ & $50.78$ & $55.11$ & $52.89$ & \cellcolor{green!25}$\mathbf{56.27}$ & $54.42$ & $53.22$ & $54.29$ & \cellcolor{green!25}$\mathbf{54.65}$ \\
Credit Approval (\href{https://archive.ics.uci.edu/dataset/27}{27}) & $690 \times 15$ & $85.01$ & $85.20$ & $81.78$ & $82.19$ & $82.24$ & \cellcolor{green!25}$\mathbf{82.85}$ & $82.17$ & \cellcolor{green!25}$\mathbf{83.03}$ & $82.93$ & $82.55$ \\
Default Credit (\href{https://archive.ics.uci.edu/dataset/350}{350}) & $30000 \times 23$ & $42.14$ & $49.36$ & $42.40$ & $50.62$ & $50.04$ & \cellcolor{green!25}$\mathbf{50.73}$ & $40.26$ & $50.15$ & $50.03$ & \cellcolor{green!25}$\mathbf{50.46}$ \\
Diabetes (\href{https://archive.ics.uci.edu/dataset/529}{529}) & $520 \times 16$ & $92.42$ & $96.15$ & $89.20$ & $92.19$ & $92.16$ & \cellcolor{green!25}$\mathbf{92.33}$ & $88.83$ & $90.57$ & $90.98$ & \cellcolor{green!25}$\mathbf{91.56}$ \\
Dry Bean (\href{https://archive.ics.uci.edu/dataset/602}{602}) & $13611 \times 16$ & $93.40$ & $93.95$ & $93.08$ & $93.67$ & \cellcolor{green!25}$\mathbf{93.70}$ & $93.63$ & $92.43$ & $93.30$ & $93.37$ & \cellcolor{green!25}$\mathbf{93.47}$ \\
German Credit (\href{https://archive.ics.uci.edu/dataset/144}{144}) & $1000 \times 20$ & $53.89$ & $60.59$ & $52.92$ & $56.60$ & $52.16$ & \cellcolor{green!25}$\mathbf{57.95}$ & $50.87$ & $57.09$ & $49.55$ & \cellcolor{green!25}$\mathbf{58.28}$ \\
Glass (\href{https://archive.ics.uci.edu/dataset/42}{42}) & $214 \times 9$ & $48.50$ & $66.82$ & $52.86$ & \cellcolor{green!25}$\mathbf{56.07}$ & $54.55$ & $55.91$ & $52.90$ & $58.29$ & $54.41$ & \cellcolor{green!25}$\mathbf{61.53}$ \\
Handwritten Digits (\href{https://archive.ics.uci.edu/dataset/80}{80}) & $5620 \times 64$ & $96.92$ & $98.63$ & $96.56$ & $98.68$ & $98.68$ & \cellcolor{green!25}$\mathbf{98.84}$ & $95.78$ & $98.03$ & $98.07$ & \cellcolor{green!25}$\mathbf{98.28}$ \\
Healthy Aging (\href{https://archive.ics.uci.edu/dataset/936}{936}) & $714 \times 14$ & $36.13$ & $33.15$ & $21.38$ & $22.94$ & $22.94$ & \cellcolor{green!25}$\mathbf{22.94}$ & $22.94$ & $22.94$ & $22.94$ & \cellcolor{green!25}$\mathbf{23.22}$ \\
Heart Failure (\href{https://archive.ics.uci.edu/dataset/519}{519}) & $299 \times 12$ & $65.94$ & $62.85$ & $67.74$ & \cellcolor{green!25}$\mathbf{69.45}$ & $63.63$ & $63.43$ & $67.59$ & $61.98$ & $64.32$ & \cellcolor{green!25}$\mathbf{70.87}$ \\
Hepatitis (\href{https://archive.ics.uci.edu/dataset/46}{46}) & $155 \times 19$ & $94.46$ & $93.75$ & $93.99$ & \cellcolor{green!25}$\mathbf{94.88}$ & $92.65$ & $94.49$ & $90.82$ & $91.99$ & $91.64$ & \cellcolor{green!25}$\mathbf{92.31}$ \\
Higher Education (\href{https://archive.ics.uci.edu/dataset/856}{856}) & $145 \times 31$ & $21.29$ & $17.47$ & $12.28$ & $9.47$ & $4.68$ & \cellcolor{green!25}$\mathbf{13.05}$ & \cellcolor{green!25}$\mathbf{11.63}$ & $8.98$ & $5.35$ & $9.28$ \\
Ionosphere (\href{https://archive.ics.uci.edu/dataset/52}{52}) & $351 \times 34$ & $91.99$ & $95.93$ & $92.11$ & $93.85$ & $95.16$ & \cellcolor{green!25}$\mathbf{95.79}$ & $92.52$ & \cellcolor{green!25}$\mathbf{96.21}$ & $95.81$ & $95.78$ \\
Iranian Churn (\href{https://archive.ics.uci.edu/dataset/563}{563}) & $3150 \times 13$ & $63.01$ & $88.44$ & $51.49$ & $80.42$ & $76.86$ & \cellcolor{green!25}$\mathbf{83.73}$ & $52.87$ & $71.86$ & $72.18$ & \cellcolor{green!25}$\mathbf{74.83}$ \\
Iris (\href{https://archive.ics.uci.edu/dataset/53}{53}) & $150 \times 4$ & $93.98$ & $96.00$ & $93.22$ & $95.99$ & \cellcolor{green!25}$\mathbf{97.33}$ & $96.66$ & $93.27$ & $94.65$ & $92.65$ & \cellcolor{green!25}$\mathbf{94.66}$ \\
Kidney Disease (\href{https://archive.ics.uci.edu/dataset/336}{336}) & $400 \times 24$ & $99.21$ & $92.47$ & $97.63$ & $97.09$ & $96.86$ & \cellcolor{green!25}$\mathbf{97.90}$ & $97.11$ & $96.84$ & $96.55$ & \cellcolor{green!25}$\mathbf{97.88}$ \\
Letter Recognition (\href{https://archive.ics.uci.edu/dataset/59}{59}) & $20000 \times 16$ & $76.87$ & $97.33$ & $75.48$ & $95.38$ & $92.30$ & \cellcolor{green!25}$\mathbf{95.88}$ & $71.73$ & $93.04$ & \cellcolor{green!25}$\mathbf{93.49}$ & $93.39$ \\
Maternal Health (\href{https://archive.ics.uci.edu/dataset/863}{863}) & $1014 \times 6$ & $63.37$ & $74.28$ & $63.13$ & $70.20$ & $68.46$ & \cellcolor{green!25}$\mathbf{70.99}$ & $63.13$ & $66.56$ & $66.23$ & \cellcolor{green!25}$\mathbf{67.80}$ \\
Mushroom (\href{https://archive.ics.uci.edu/dataset/73}{73}) & $8124 \times 22$ & $99.96$ & $99.96$ & $99.44$ & $99.94$ & $99.92$ & \cellcolor{green!25}$\mathbf{99.97}$ & $98.95$ & $99.82$ & $99.86$ & \cellcolor{green!25}$\mathbf{99.91}$ \\
Obesity (\href{https://archive.ics.uci.edu/dataset/544}{544}) & $2111 \times 16$ & $90.91$ & $96.59$ & $81.97$ & $91.86$ & $90.74$ & \cellcolor{green!25}$\mathbf{92.48}$ & $74.64$ & $88.77$ & $90.01$ & \cellcolor{green!25}$\mathbf{91.70}$ \\
Online Shopping (\href{https://archive.ics.uci.edu/dataset/468}{468}) & $12330 \times 17$ & $53.51$ & $66.10$ & $52.43$ & $64.75$ & $64.92$ & \cellcolor{green!25}$\mathbf{66.04}$ & $51.76$ & $60.91$ & $59.91$ & \cellcolor{green!25}$\mathbf{61.55}$ \\
Rice (\href{https://archive.ics.uci.edu/dataset/545}{545}) & $3810 \times 7$ & $92.97$ & $93.07$ & $93.06$ & $93.98$ & $94.05$ & \cellcolor{green!25}$\mathbf{94.21}$ & $93.11$ & $93.96$ & $93.75$ & \cellcolor{green!25}$\mathbf{94.05}$ \\
Seoul Bike Sharing (\href{https://archive.ics.uci.edu/dataset/560}{560}) & $8760 \times 13$ & $99.76$ & $99.88$ & $99.70$ & \cellcolor{green!25}$\mathbf{99.89}$ & $99.86$ & \cellcolor{green!25}$\mathbf{99.89}$ & $99.68$ & $99.75$ & $99.70$ & \cellcolor{green!25}$\mathbf{99.76}$ \\
Spambase (\href{https://archive.ics.uci.edu/dataset/94}{94}) & $4601 \times 57$ & $90.12$ & $91.58$ & $90.22$ & $91.96$ & $90.99$ & \cellcolor{green!25}$\mathbf{92.18}$ & $89.22$ & $89.34$ & $89.83$ & \cellcolor{green!25}$\mathbf{90.07}$ \\
Student Dropout (\href{https://archive.ics.uci.edu/dataset/697}{697}) & $4424 \times 36$ & $68.14$ & $68.22$ & $66.09$ & $67.62$ & $68.73$ & \cellcolor{green!25}$\mathbf{68.85}$ & $63.94$ & $65.78$ & $66.15$ & \cellcolor{green!25}$\mathbf{67.22}$ \\
Telescope (\href{https://archive.ics.uci.edu/dataset/159}{159}) & $19020 \times 10$ & $66.62$ & $81.70$ & $66.69$ & $80.65$ & $80.67$ & \cellcolor{green!25}$\mathbf{81.17}$ & $67.17$ & $79.69$ & $79.18$ & \cellcolor{green!25}$\mathbf{80.32}$ \\
Wholesale (\href{https://archive.ics.uci.edu/dataset/292}{292}) & $440 \times 7$ & $26.64$ & $29.30$ & $27.60$ & $27.65$ & $27.81$ & \cellcolor{green!25}$\mathbf{28.72}$ & $27.40$ & $29.03$ & $27.81$ & \cellcolor{green!25}$\mathbf{31.20}$ \\
Wine (\href{https://archive.ics.uci.edu/dataset/109}{109}) & $178 \times 13$ & $97.34$ & $96.71$ & \cellcolor{green!25}$\mathbf{97.34}$ & $96.13$ & $94.97$ & $97.27$ & \cellcolor{green!25}$\mathbf{97.27}$ & $95.01$ & $95.69$ & $96.13$ \\
\midrule
Average &  & $73.60$ & $78.76$ & $71.44$ & $75.96$ & $74.95$ & \cellcolor{green!25}$\mathbf{76.68}$ & $70.66$ & $74.93$ & $74.32$ & \cellcolor{green!25}$\mathbf{76.51}$ \\
Relative Improvement &  &  &  &  &  & $-1.54$ & \cellcolor{green!25}$\mathbf{+0.99}$ &  &  & $-0.79$ & \cellcolor{green!25}$\mathbf{+1.90}$ \\
\bottomrule
\end{tabularx}
\end{sc}
\end{small}
\end{center}
\end{table*}

%% file: tab_reg.tex
\begin{table*}[t]
\caption{Test R$^2$ average over 5 bootstrap executions for each regression dataset. The best model is highlighted in \texthl{green}{\textbf{bold}}.}
\label{tab:regression_results}
\begin{center}
\begin{small}
\begin{sc}
\begin{tabularx}{\textwidth}{@{} l c | Y Y | Y Y Y c | Y Y Y c}
\toprule
Model; Dataset (\href{https://archive.ics.uci.edu/}{UCI ID}) & Shape & \multicolumn{2}{c|}{Clean} & \multicolumn{4}{c|}{CCAR} & \multicolumn{4}{c}{CNAR} \\
\midrule
Linear &  & $\checkmark$ &  & $\checkmark$ &  &  &  & $\checkmark$ &  &  &  \\
MLP &  &  & $\checkmark$ &  & $\checkmark$ & $\checkmark$ & $\checkmark$ &  & $\checkmark$ & $\checkmark$ & $\checkmark$ \\
Curriculum &  &  &  &  &  & $\checkmark$ & ($\checkmark$) &  &  & $\checkmark$ & ($\checkmark$) \\
QuAIL &  &  &  &  &  &  & $\checkmark$ &  &  &  & $\checkmark$ \\
\midrule
Abalone (\href{https://archive.ics.uci.edu/dataset/1}{1}) & $4177 \times 8$ & $54.31$ & $63.26$ & $50.90$ & $56.31$ & $55.63$ & \cellcolor{green!25}$\mathbf{56.84}$ & $48.12$ & $54.57$ & \cellcolor{green!25}$\mathbf{55.12}$ & $55.09$ \\
Appliance Energy (\href{https://archive.ics.uci.edu/dataset/374}{374}) & $19735 \times 28$ & $15.26$ & $40.65$ & $13.84$ & $31.01$ & $19.41$ & \cellcolor{green!25}$\mathbf{32.37}$ & $11.59$ & $23.34$ & $20.61$ & \cellcolor{green!25}$\mathbf{23.35}$ \\
Auto MPG (\href{https://archive.ics.uci.edu/dataset/9}{9}) & $398 \times 7$ & $82.70$ & $88.50$ & $82.84$ & $88.14$ & $82.47$ & \cellcolor{green!25}$\mathbf{88.19}$ & $81.32$ & $83.61$ & $86.74$ & \cellcolor{green!25}$\mathbf{87.04}$ \\
Automobile (\href{https://archive.ics.uci.edu/dataset/10}{10}) & $205 \times 25$ & $59.17$ & $67.42$ & $49.23$ & $59.15$ & $58.55$ & \cellcolor{green!25}$\mathbf{66.10}$ & $46.08$ & $56.68$ & $53.29$ & \cellcolor{green!25}$\mathbf{57.12}$ \\
Banknote Auth (\href{https://archive.ics.uci.edu/dataset/267}{267}) & $1372 \times 4$ & $85.15$ & $99.94$ & $83.78$ & $98.34$ & $91.75$ & \cellcolor{green!25}$\mathbf{98.98}$ & $80.52$ & $98.22$ & $95.59$ & \cellcolor{green!25}$\mathbf{98.49}$ \\
Bike Sharing (\href{https://archive.ics.uci.edu/dataset/275}{275}) & $17379 \times 13$ & $68.18$ & $92.79$ & $-32.45$ & \cellcolor{green!25}$\mathbf{27.35}$ & $27.20$ & $27.34$ & $-33.75$ & $25.24$ & $23.91$ & \cellcolor{green!25}$\mathbf{26.29}$ \\
Concrete (\href{https://archive.ics.uci.edu/dataset/165}{165}) & $1030 \times 8$ & $61.12$ & $88.40$ & $59.22$ & $85.45$ & $84.83$ & \cellcolor{green!25}$\mathbf{86.42}$ & $57.94$ & $82.75$ & \cellcolor{green!25}$\mathbf{83.30}$ & $82.87$ \\
Garment (\href{https://archive.ics.uci.edu/dataset/597}{597}) & $1197 \times 14$ & $30.06$ & $33.69$ & $29.03$ & $35.39$ & $27.53$ & \cellcolor{green!25}$\mathbf{35.83}$ & $28.32$ & $33.46$ & $24.97$ & \cellcolor{green!25}$\mathbf{33.64}$ \\
Heart Disease (\href{https://archive.ics.uci.edu/dataset/45}{45}) & $303 \times 13$ & $45.34$ & $49.55$ & $10.59$ & $44.54$ & $34.78$ & \cellcolor{green!25}$\mathbf{47.58}$ & $12.30$ & $46.93$ & $35.45$ & \cellcolor{green!25}$\mathbf{48.85}$ \\
Liver Disorders (\href{https://archive.ics.uci.edu/dataset/60}{60}) & $345 \times 5$ & $12.38$ & $7.13$ & \cellcolor{green!25}$\mathbf{13.72}$ & $9.56$ & $10.01$ & $12.51$ & $6.75$ & $-1.31$ & $15.56$ & \cellcolor{green!25}$\mathbf{17.67}$ \\
Parkinson (\href{https://archive.ics.uci.edu/dataset/174}{174}) & $195 \times 22$ & $42.66$ & $80.34$ & $37.40$ & $64.86$ & $54.50$ & \cellcolor{green!25}$\mathbf{69.59}$ & $31.06$ & $49.29$ & \cellcolor{green!25}$\mathbf{59.89}$ & $51.95$ \\
Phishing Websites (\href{https://archive.ics.uci.edu/dataset/327}{327}) & $11055 \times 30$ & $70.53$ & $90.06$ & $70.40$ & \cellcolor{green!25}$\mathbf{89.38}$ & $88.59$ & $89.24$ & $69.91$ & $85.61$ & $85.48$ & \cellcolor{green!25}$\mathbf{85.81}$ \\
Power Plant (\href{https://archive.ics.uci.edu/dataset/294}{294}) & $9568 \times 4$ & $92.88$ & $94.11$ & $91.08$ & $92.16$ & $91.65$ & \cellcolor{green!25}$\mathbf{92.62}$ & $89.86$ & $92.96$ & $93.03$ & \cellcolor{green!25}$\mathbf{93.25}$ \\
Real Estate (\href{https://archive.ics.uci.edu/dataset/477}{477}) & $414 \times 6$ & $52.47$ & $60.86$ & $52.29$ & $63.90$ & $58.16$ & \cellcolor{green!25}$\mathbf{65.41}$ & $52.28$ & $63.71$ & $63.22$ & \cellcolor{green!25}$\mathbf{65.27}$ \\
Wine Quality (\href{https://archive.ics.uci.edu/dataset/186}{186}) & $6497 \times 11$ & $29.48$ & $41.13$ & $29.34$ & $37.97$ & $38.02$ & \cellcolor{green!25}$\mathbf{38.65}$ & $28.74$ & \cellcolor{green!25}$\mathbf{38.09}$ & $37.23$ & $37.65$ \\
\midrule
Average &  & $53.57$ & $68.52$ & $44.81$ & $59.66$ & $55.49$ & \cellcolor{green!25}$\mathbf{61.24}$ & $42.69$ & $56.63$ & $55.56$ & \cellcolor{green!25}$\mathbf{57.55}$ \\
Relative Improvement &  &  &  &  &  & $-5.63$ & \cellcolor{green!25}$\mathbf{+2.99}$ &  &  & $-5.63$ & \cellcolor{green!25}$\mathbf{+1.30}$ \\
\bottomrule
\end{tabularx}
\end{sc}
\end{small}
\end{center}
\end{table*}

%% file: tab_comp.tex
\begin{table}[t]
\caption{Average relative improvement of QuAIL with respect to MLP.}
\label{tab:comparison}
\begin{center}
\begin{small}
\begin{sc}
\begin{tabularx}{\linewidth}{@{} l l Y c @{}}
\toprule
Stratification & Category & Count & Impr. (\%) \\
\midrule
& Small & 22 & 2.95 \\
Dataset Size & Medium & 18 & 1.06 \\
& Large & 10 & 0.72 \\
\midrule
& Low & 17 & 2.62 \\
N. Features & Medium & 17 & 1.25 \\
& High & 16 & 1.57 \\
\midrule
& Mostly Linear & 17 & 1.69 \\
Linearity & Mixed & 16 & 2.03 \\
& Mostly Non-Linear & 17 & 1.82 \\
\bottomrule
\end{tabularx}
\end{sc}
\end{small}
\end{center}
\end{table}